\documentclass[letterpaper]{article} % DO NOT CHANGE THIS
\usepackage{aaai25}  % DO NOT CHANGE THIS
\usepackage{times}  % DO NOT CHANGE THIS
\usepackage{helvet}  % DO NOT CHANGE THIS
\usepackage{courier}  % DO NOT CHANGE THIS
\usepackage[hyphens]{url}  % DO NOT CHANGE THIS
\usepackage{graphicx} % DO NOT CHANGE THIS
\usepackage{natbib}  % DO NOT CHANGE THIS AND DO NOT ADD ANY OPTIONS TO IT
\usepackage{caption} % DO NOT CHANGE THIS AND DO NOT ADD ANY OPTIONS TO IT
\usepackage{algorithm}
\usepackage{algorithmic}
\usepackage{amsmath}    %, bbm

\usepackage{newfloat}
\usepackage{listings}
\DeclareCaptionStyle{ruled}{labelfont=normalfont,labelsep=colon,strut=off} % DO NOT CHANGE THIS
\floatstyle{ruled}
\newfloat{listing}{tb}{lst}{}
\floatname{listing}{Listing}
\usepackage{booktabs}   %user added
\usepackage{fontawesome}   %user added
\usepackage{amsmath}    %user added
\usepackage{makecell,multirow,diagbox}  %user added
\usepackage{color}  %user added

\title{Enhancing Trustworthiness of Graph Neural Networks with Rank-Based Conformal Training}
\author {
    Ting Wang\textsuperscript{\rm 1},
    Zhixin Zhou\textsuperscript{\rm 2},
    Rui Luo\textsuperscript{\rm 1}\thanks{Corresponding author: } 
}
\affiliations {
    \textsuperscript{\rm 1}Department of Systems Engineering, City University of Hong Kong, Hong Kong SAR, China\\
    \textsuperscript{\rm 2}Alpha Benito Research, Los Angeles, USA\\
    t.wang123@outlook.com, zzhou@alphabenito.com, ruiluo@cityu.edu.hk
}
\usepackage{bibentry}
\begin{document}

\maketitle

\begin{abstract}
Graph Neural Networks (GNNs) has been widely used in a variety of fields because of their great potential in representing graph-structured data. However, lacking of rigorous uncertainty estimations limits their application in high-stakes. Conformal Prediction (CP) can produce statistically guaranteed uncertainty estimates by using the classifier's probability estimates to obtain prediction sets, which contains the true class with a user-specified probability. In this paper, we propose a Rank-based CP during training framework to GNNs (RCP-GNN) for reliable uncertainty estimates to enhance the trustworthiness of GNNs in the node classification scenario. By exploiting rank information of the classifier's outcome, prediction sets with desired coverage rate can be efficiently constructed. The strategy of CP during training with differentiable rank-based conformity loss function is further explored to adapt prediction sets according to network topology information. In this way, the composition of prediction sets can be guided by the goal of jointly reducing inefficiency and probability estimation errors. Extensive experiments on several real-world datasets show that our model achieves any pre-defined target marginal coverage while significantly reducing the inefficiency compared with state-of-the-art methods.
\end{abstract}

\begin{links}
    \link{Code}{https://github.com/CityU-T/RCP-GNN}
\end{links}

\section{Introduction}

Graph Neural Networks (GNNs) has been widely used in many applications, such as weather forecasting \cite{doi:10.1126/science.adi2336}, drug discovery \cite{GRLBH} and recommendation systems \cite{10.1145/3535101}. 
However, predictions made by GNNs are inevitably present uncertainty. Though to understand the uncertainty in the predictions they produce can help to enhance the trustworthiness of GNNs \cite{NEURIPS2023_54a1495b}, most existing uncertainty quantification (UQ) methods can not be easily adopted to graph-structured data \cite{NEURIPS2021_95b431e5}. Among various UQ methods, Conformal Prediction (CP) is an effective approach for achieving trustworthy GNNs \cite{wang2024uncertainty}.  It relaxes the assumption of existing UQ methods, making it suitable for graph-structured data.

Instead of relying solely on uncalibrated predicted distribution \(\mu(y|x)\), CP constructs a prediction set that informs a plausible range of estimates aligned with the true outcome distribution \(p(y|x)\). The post-training calibration step make the output prediction sets provably include the true outcome with a user-specified coverage of \(1-\alpha\). A conformity score function is the key component for CP to quantify the agreement between the input and the candidate label.

Some studies have draw attention to the application of CP to graph-structured data \cite{pmlr-v202-clarkson23a, pmlr-v202-h-zargarbashi23a, lunde2023validityconformalpredictionnetwork, lunde2023conformalpredictionnetworkassistedregression, marandon2024conformallinkpredictionfalse}. 
However, CP usually suffer inefficiency when there are no well-calibrated probabilities, with the intuition that larger prediction sets covers higher uncertainty. How to achieve desirable efficiency beyond validity is still a noteworthy challenge. Existing studies \cite{Sadinle_2018, NEURIPS2020_244edd7e} are typically changing the definition of the conformity scores for inefficiency reduction. The challenge is that CP is always used as a post-training calibration step, thus hindering its ability of underlying model to adapting to the prediction sets. 

Recently, \cite{stutz2022learning, Bellotti2021OptimizedCC} try to using CP as a training step to make model parameter \(\theta\) dependent with the calibration step, so as to modifying prediction sets towards reducing inefficiency.
However, the integration of conformal training and GNNs still remain largely unexplored. The very resent work \cite{NEURIPS2023_54a1495b} proposed a conformal graph neural network which develops a topology-aware calibration step during training.
% Node classification task
Differently, we focus this problem with two lines. One is that a suitable conformity score is applied for GNNs who often struggle with miscalibration predictions \cite{NEURIPS2021_c7a9f13a}. Another is that a conformal training framework based on the differentiable variant of this conformity score is designed to adjust the prediction sets along with model parameters' optimization.

In conclusion, our contributions are two-fold. First, we propose a novel rank-based conformity scores that emphasizes the rank of prediction probabilities which is more robust to GNNs. Second, we develops a calibration step during training for adjusting the prediction sets along with the model parameters. We demonstrate that the rank-based conformal prediction method we introduce is performance-critical for efficiently constructing prediction sets with expected coverage rate. And the proposed method can outperform state-of-the-art methods for the graph node classification tasks on several popular network datasets in terms of the converge and inefficiency metrics.

\section{Preliminaries}

Let \(G=(\mathcal{V}, \mathcal{E}, X)\) be a graph, where \(\mathcal{V}\) is a set of nodes, \(\mathcal{E}\) is a set of edges, and  \(X = \{ x_v \}_{v \in \mathcal{V}}\) is the attributes. We denote \( \mathcal{Y}\) as the discrete set of possible label classes. Let \( \{(x_v, y_v)\}_{v \in D}\) be the random variables from the training data, where \(x_v \in R^d\) is the \(d\)-dimensional vector for node \(v\) and \(y_v \in \mathcal{Y}\) is its corresponding class. The training data \(\mathcal{D}\) is randomly split into  \(\mathcal{D}_{tr}\)/\(\mathcal{D}_{val}\)/\(\mathcal{D}_{calib}\) as training/validation/calibration set. Note that the subset \(\mathcal{D}_{calib}\) is withhold as calibration data for conformal prediction. Let \( \{(x_v)\}_{v \in D_{te}}\) be the random variables from the test data whose true labels \( \{(y_v)\}_{v \in D_{te}}\) is unknown for model. The goal of node classification tasks is to obtain a classifier \(\mu: X \rightarrow Y\), which can approximate the posterior distribution over classes \(y_v \in Y\). During the training step, \( \{(x_v, y_v)\}_{v \in D_{tr} \cup D_{valid}}\), \( \{(x_v)\}_{v \in D_{te } \cup D_{calib}}\) and the graph structure \((\mathcal{V}, \mathcal{E})\) are available to GNN model for node representations.

\subsection{Graph Neural Networks (GNNs)}
In this paper, we focus on GNNs in the node classification scenario. GNNs is the most common encoder to learn compact node representations, which is generated by a series of propagation layers. For each layer \(l\), each node representation \(h_u^{(l)}\) is updated by its previous representations \(h_u^{(l-1)}\), and aggregated features \(\hat{m}_u^{(l)}\) obtained through passing message from its neighbours \(\mathcal{N}_{(u)}\):
\begin{equation}
h_u^{(l)} =\textrm{F}_\textrm{upd}(h_u^{(l-1)}, \hat{m}_u^{(l)})\\
\end{equation}
\begin{equation}
\hat{m}_u^{(l)} = \textrm{F}_\textrm{agg}(m_{(uv)},  | v \in \mathcal{N}_{(u)}) \\
\end{equation}
\begin{equation}
m_{(uv)}=\textrm{F}_\textrm{msg}(h_u^{(l-1)}, h_v^{(l-1)}) 
\end{equation}
where \(\textrm{F}_\textrm{upd}(\cdot)\) is a non-linear function to update node representations. \(\textrm{F}_\textrm{agg}(\cdot)\) is the aggregation function while \(\textrm{F}_\textrm{msg}(\cdot)\) is the message passing function. We use node representations in the last layer as the input of a classifier to obtain a prediction \(\mu_{\theta}(x)\).

For CP on GNNs, a valid coverage guarantee requires the exchangeability of the calibration and test data. Since our model is transduction node classification, the calibration examples are drawn exchangeability from the test distribution following \cite{NEURIPS2023_54a1495b}.

\subsection{Conformal Prediction}

For a new test point \(x_{n+1}\), the goal of CP is to construct a reasonably small prediction set \(C(x_{n+1})\), which contains corresponding true label \(y_{n+1} \in \mathcal{Y}\) with pre-defined coverage rate \(1-\alpha\):
\begin{equation}
    P(y_{n+1} \in C(x_{n+1})) \geq  1 - \alpha
\end{equation}
where \(\alpha \in [0, 1]\) is the user-specific miscoverage rate. The standard CP is usually conduct at test time after the classification model \(\mu_{\theta}\) is trained, which is achieved in two steps: 1) \textit{In the calibration step}, a cut-off threshold \(\hat{\eta}\) is calculated by the quantile function of the conformity scores  \(V:X \times Y \rightarrow R\) on the hold-out calibration set \(\mathcal{D}_{calib}\). During calibration, the true classes \(y_i\) are used for computing the threshold to ensure coverage \(1 - \alpha\).
2)  \textit{In the prediction step}, the prediction sets \(C(x)\) depending on the threshold \(\hat{\eta}\) and the model parameters \(\theta\) are constructed.
The conformity score function is designed to measure the predicted probability of a class, and it is typically changed for various objectives. Two popular conformity scores are described in details below.
 
\subsubsection{Threshold Prediction Set (THR)}
 The threshold \(\hat{\eta}\) for THR \cite{Sadinle_2018} is calculated by the \(\alpha\) quantile of the conformity scores:
\begin{equation}
    \hat{\eta} = Q(\{V (x_i, y_j)|i \in \mathcal{D}_{calib}\}, \alpha(1 + \frac{1}{|\mathcal{D}_{calib}|}))
\end{equation}
where \(Q(\cdot)\) is the quantile function. The prediction sets including labels with sufficiently large prediction values are constructed by thresholding probabilities: 
\begin{equation}
 C(x) = \{k \in \mathcal{Y}: V (x, k) \geq \hat{\eta}\}
 \end{equation}
 \begin{equation}
 \label{thr_v}
 V (x, k) =  \mu_k(x)
\end{equation}

\subsubsection{Adaptive Prediction Set (APS)}
APS \cite{NEURIPS2020_244edd7e} takes the cumulative sum of ordered probabilities \(\mu_{\pi(1)}(x) > \mu_{\pi(2)}(x) > \cdots \mu_{\pi(|\mathcal{Y}|)}(x)\) for prediction set construction:
\begin{equation}
C(x) = \{k \in \mathcal{Y}: V (x, k) \leq \hat{\eta}\}
\end{equation}
    \begin{equation}
    \label{aps_v}
    V (x, k) = \sum_{j=1}^{k}\mu_{\pi(j)}(x)
\end{equation}
where \(\pi\) is a permutation of \(\mathcal{Y}\), and the \((1-\alpha)(1 + 1/|\mathcal{D}_{calib}| )\)-quantile is also required for calibration to ensure marginal coverage.

\section{RANK: Rank-based Conformal Prediction}

Following our previous work \cite{luo2024trustworthy}, we advancing CP to GNNs through rank-based conformity scores, named RANK, to directly reduce the inefficiency. Assuming that a higher value of \(\mu_k(x_i)\) indicates a greater likelihood of \(x_i\) belonging to class \(k\). Consequently, if class \(k\) is included in the prediction set, and \(\mu_{k'}(x_i)>\mu_{k}(x_i)\) satisfied, then \(k'\) must be in the prediction set. According to this assumption, the size of the prediction set including the true label can be evaluated in the calibration step. The smallest prediction set that includes the true label \(y_i\) will be constructed by the rank of \(\mu_{y_i}(x_i)\) within the sequence \(\{\mu_{1}(x) ,\cdots \mu_{\mathcal{Y}}(x)\}\).

\subsubsection{Ranked Threshold Prediction Sets}

For each \(i \in \mathcal{D}_{calib}\), the following rank is defined to establish a rule to choose labels:
\begin{equation} 
r_i = \mathrm{rank} \;\mathrm{of} \; \mu_{y_i}(x_i) \; \mathrm{in} \; \{\mu_k(x_i): k \in \mathcal{Y}\}
\end{equation}
so that the order statistics can be find: \(r_{(1)} \geq r_{(2)}  \cdots \geq r_{(n)}\). Let \(r_{\alpha}^{*} = r_{(\lfloor (n+1)\alpha \rfloor)}\), either top-\((r_{\alpha}^{*}-1)\) or top-\((r_{\alpha}^{*})\) classes will be included in the prediction set. The top-\((r_{\alpha}^{*})\) classes refers to the classes corresponding to the \((r_{\alpha}^{*})\)-th largest prediction values. To achieve the target coverage, the \(\mu^*\) is defined to determine when the \((r_{\alpha}^{*})\)-th class should be included in the prediction sets:
\begin{equation} 
\mu^* = \lceil np \rceil\textrm{-th largest value in } \{\hat{\mu}_{r_{\alpha}^{*}}(x_i): i \in \mathcal{D}_{calib}\}
\end{equation}
where \(n\) is the size of \(\mathcal{D}_{calib}\), \(p\) is the proportion of instances we should included in the \(r_a^*\)-th label, and \(\hat{\mu}_{k}(x_i)\) denotes the \(k\)-th order statistics in \((\mu_{1}(x_i),...,\mu_{n}(x_i))\). The prediction set is defined as follows:
\begin{equation} 
\label{rankedC}
C(x) = \left \{
\begin{array}{ll}
     & \{ k \in \mathcal{Y}: \hat{\mu}_k(x) \geq \hat{\mu}_{r_{\alpha}^{*}}(x) \},  \\
     & \textrm{if} \quad \hat{\mu}_{r_{\alpha}^{*}}(x) \geq \mu^* ;\\
     & \{ k \in \mathcal{Y}: \hat{\mu}_k(x) \geq \hat{\mu}_{r_{{\alpha}^{*} - 1}}(x) \},  \\
     & \textrm{otherwise};\\
     \end{array}
     \right.
\end{equation}
According to above analysis, the rank-based conformity scores calculated on the calibration set can be defined following:
\begin{equation}
    \label{rank_v}
    \begin{split}
    & V (x_i, y_i) = \\
    &[\mathrm{rank} \;\mathrm{of} \; \hat{\mu}_{y_i}(x_i) \; \mathrm{in} \; \{\hat{\mu}_1(x_i), \cdots, \hat{\mu}_\mathcal{Y}(x_i)\}] - 1\\
    & + \frac{1}{n} [\mathrm{rank} \;\mathrm{of} \; \hat{\mu}_{y_i}(x_i) \; \mathrm{in} \; \{\hat{\mu}_{y_i}(x_1), \cdots, \hat{\mu}_{y_i}(x_n)\}]
    \end{split}
\end{equation}
and the quantile \(Q\) as the \(\lfloor (n+1)\alpha \rfloor\)-th largest value among the conformity scores, defining the prediction set is equivalent to selecting the calibration samples that satisfy the condition \(V(x_i,y_i) \leq Q\), is employed to construct the prediction set with \(1-\alpha\) coverage.
% where \( \hat{\mu}_k(x)\) denotes the \(k\)-th order statistics in \( (\mu_{1}(x) ,\cdots \mu_{n}(x) )\).
% Then, using eq.~(\ref{rankedC})

\section{RCP-GNN: Rank-Based Conformal Prediction on Graph Neural Networks}

\begin{figure*}[t]
\centering
\includegraphics[width=1\textwidth]{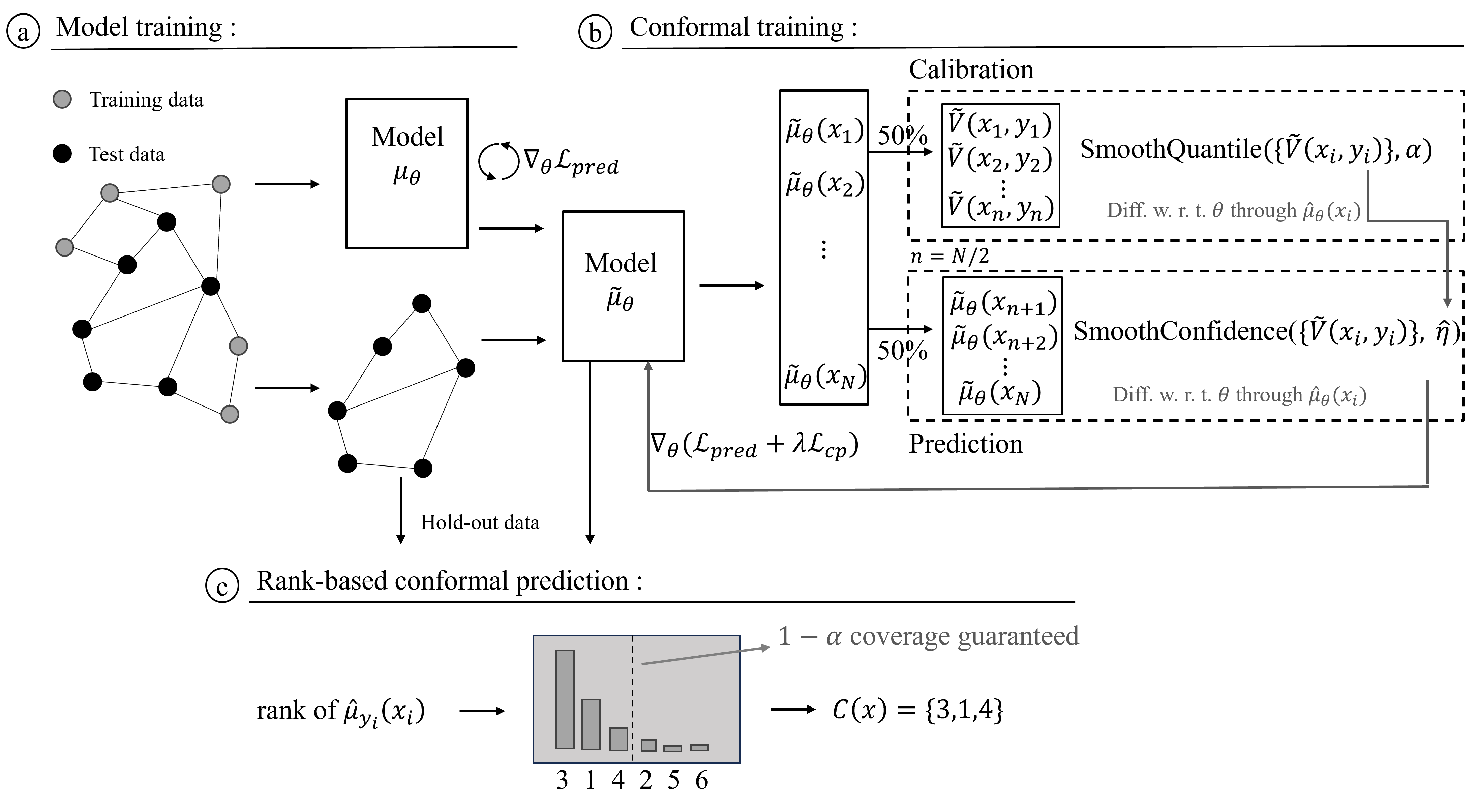} % Reduce the figure size so that it is slightly narrower than the column. Don't use precise values for figure width.This setup will avoid overfull boxes.
\caption{The framework of RCP-GNN. (a) \underline{Model training stage.} The GNN model \(\textrm{GNN}_{base}\) is trained by optimizing a prediction loss using a standard deep learning step. And the prediction probabilities of node \(i\): \(\mu(x_i)\) is obtained. (b) \underline{Conformal training stage.} The novel rank-based conformal training step is proposed to adjust the prediction set for desirable properties jointly with improve estimation accuracy. The topology-aware correction model \(\textrm{GNN}_{cor}\) that takes \(\tilde{\mu}(x)\) as the output is updated by the conformal training step.
% The differentiable conformity loss is optimizing along with the prediction loss. 
(c) \underline{Rank-based Conformal Prediction.} The rank-based CP is employed to produce a prediction set based on \(\tilde{\mu}\) which includes true label with a user-specified probability.}
\label{fig1}
\end{figure*}

We propose RCP-GNN into two-stage: model training stage and conformal training stage, as Figure \ref{fig1} shows. In model training stage, the base model \(\textrm{GNN}_{base}\) is trained only by prediction loss (i.e., cross-entropy loss), and \(\mu(X)\) is the estimator of base model.
% and standard rank-based conformal predictors are applied at test time.

In conformal training stage, the correct model \(\textrm{GNN}_{cor}\) is trained by both prediction loss and conformity loss. We set \(\tilde{\mu}(X) = \mathrm{GNN}_{cor}(\mu(X), G)\) as the estimator of correct model. The prediction set will be optimized when CP performing on each mini-batch. For the reason that training with original rank-based CP may cause limited gradient flow, a differentiable implementation for RANK is designed to enable smooth sorting and ranking, and is further used to construct the conformity loss function, which sharpens the prediction set.

\subsection{Conformal Training}
We try to train our model end-to-end with the conformal wrapper in order to allow fine-grained control over the prediction sets \(C(x)\). Following the \textit{split} CP approach \cite{Lei2013ACP}, we randomly split the test data set \(\mathcal{D}_{te}\) into folds with 50\%/50\% as  \(\hat{\mathcal{D}}_{calib}\)/\(\hat{\mathcal{D}}_{te}\) for calibration and constructing prediction sets. Before splitting the test data, a fraction of test data is withhold for further standard rank-based conformal prediction stage.

\subsubsection{Differentiable Prediction and Calibration Steps}
A differentiable CP method which involves differentiable prediction and calibration step is defined for the training process: 1)\textit{ In the prediction step,} the prediction sets \(C(x)\) w.r.t. the threshold \(\hat{\eta}\) and the predictions \(\tilde{\mu}_\theta(x)\) is set to be differentiable. 2) \textit{In the calibration step,} the conformity scores w.r.t. the predictions \(\tilde{\mu}_\theta(x)\) as well as quantile function is set to be differentiable. Notably, the predictions \(\tilde{\mu}_\theta(x)\) are always differentiable throughout calibration and prediction steps. Therefore, The key component of differentiating through CP is the differentiable conformity scores and the differentiable quantile computation.

Given the prediction probabilities \(\tilde{\mu}(X)\), the smooth sorting designed by a \(\mathrm{sigmoid}(x) = \frac{1}{1+e^{-x}}\) function and a temperature hyper-parameter \(\tau \in [0, 1]\) is utilized to replace ``hard" rank manipulation for the smoothed rank-based conformity scores: 
\begin{equation}
\label{Rank_diff_score}
    \tilde{V} (x_i, k) = \sum_{j=1}^{|\mathcal{Y}|}\textrm{sigmoid}(\frac{ \tilde{\mu}_{j}(x_i) - \tilde{\mu}_{k}(x_i)}{\tau})
\end{equation}
After that, a differentiable quantile computation is employed for smoothed thresholding under smooth sorting.
\begin{equation}
    \hat{\eta} = \tilde{Q}(\{V( x_i, y_i ) | i \in \hat{\mathcal{D}}_{calib}\}, \alpha(1+\frac{1}{|\hat{\mathcal{D}}_{calib}|}))
\end{equation}
where \(\tilde{Q}(\cdot)\) is the smooth quantile function that are well-established in \cite{pmlr-v119-blondel20a, Chernozhukov2007QUANTILEAP}.

\subsubsection{Loss Function}

The conformal training stage performs differentiable CP on data batch during stochastic gradient descent (SGD) training. As mentioned above, the \(\hat{\eta}\) is calibrated by \(\alpha(1+1/{\hat{|\mathcal{D}}_{calib}|})\)-quantile of the conformity scores in a differentiable way. Under the constraint of hyper-parameter \(\tau\), we empirically make coverage close to \(1-\alpha\) by approximating ``hard'' sorting. Then we propose a conformity loss function to further optimize the inefficiency through training.
Given the estimator \(\tilde{\mu}_{j}(x_i)\) for the conditional probability of \(Y\) being class \(k \in \mathcal{Y}\) at \(X = x_i\) and the true label \(y_i\). Similar with Eq.\ref{Rank_diff_score}, the smooth conformity scores on test data is defined as:
\begin{equation}
    \tilde{V} (x_i, y_i) = \sum_{k=1}^{|\mathcal{Y}|}\textrm{sigmoid}(\frac{ \tilde{\mu}_{k}(x_i) - \tilde{\mu}_{y_i}(x_i)}{\tau})
\end{equation}
Given \(i \in \hat{\mathcal{D}}_{calib}\), a soft assignment \cite{stutz2022learning, NEURIPS2023_54a1495b} of each class \(k\) to the prediction set is defined smoothly as follows:
\begin{equation}
    c_{i} = \textrm{max}(0, \sum_{k \in \mathcal{Y}}{\textrm{sigmoid}(\frac{ \tilde{V} (x_i, k) - \hat{\eta}}{\tau})} - \kappa)
\end{equation}
Then the conformity loss function is defined by:
\begin{equation}
    \mathcal{L}_{cp} = \frac{1}{|\hat{\mathcal{D}}_{te}|}\frac{1}{|\mathcal{Y}|}\sum\nolimits_{i \in \hat{\mathcal{D}}_{te}}{c_i}
\end{equation}
% For APS and RANK, \(V (x_i, k)\) is the sum without order.
Thus, the loss function optimized in conformal training stage is defined as follows:
\begin{equation}
\label{eq_loss}
    \mathcal{L} = \mathcal{L}_{pred}+\lambda*\mathcal{L}_{cp}
\end{equation}
where \(\lambda\) is a hyper-parameter to balance the items and \( \mathcal{L}_{pred} \) is the prediction loss for optimizing model parameters \( \theta\):
\begin{equation}
\begin{split}
   & \mathcal{L}_{pred}  =\\
   &\!-\! \sum_{i \in \mathcal{D}_{tr}}[y_i log(\tilde{\mu}_\theta(x_i))\! +\! (1 \! - \! y_i)log(1 \!-\! \tilde{\mu}_\theta(x_i))]
\end{split}
\end{equation}
After training, standard rank-based CP are conduct on \(\tilde{\mu}(X) \) for prediction sets construction.

\section{Experiment}
We conduct experiments to demonstrate the advantages of our model over existing methods in achieving empirical marginal coverage for graph data, as well as the efficiency improvement. We also conduct systematic ablation and parameter analysis to show the robustness of our model.

\subsection{Experiment Setup}
\subsubsection{Dataset}
We choose eight popular graph-structured datasets, i.e., Cora, DBLP, CiteSeer and PubMed \cite{pmlr-v48-yanga16}, Amazon-Computers and Amazon-Photo, Coauthor-CS and Coauthor-Physics \cite{DBLP:journals/corr/abs-1811-05868} for evaluation. We randomly split them with 20\%/10\%/70\% as training/validation/testing set following previous works \cite{NEURIPS2023_54a1495b, stutz2022learning}. The statistical information of datasets is summarized in Table \ref{tab:table_a1}.
\begin{table}[tb]
\centering
\setlength{\tabcolsep}{1mm}{
\begin{tabular}{ccccc}
\toprule
Data & \# Nodes& \# Edges& \# Features& \# Labels \\
\midrule
Cora& 2,995& 16,346& 2,879& 7\\
DBLP& 17,716& 105,734&1,639& 4\\
CiteSeer& 4,230& 10,674& 602& 6\\
PubMed& 19,717& 88,648& 500& 3\\
Computers&13,752& 491,722& 767& 10\\
Photos&7,650& 238,162& 745& 8 \\
CS&18,333& 163,788& 6,805& 15 \\
Physics&34,493& 495,924& 8,415& 5 \\
\bottomrule
\end{tabular}}
\caption{Statistics of Datasets.}
\label{tab:table_a1}
\end{table}
\subsubsection{Baseline}
We consider both general statistical calibration approaches temperate, i.e., temperate scaling \cite{pmlr-v70-guo17a}, vector scaling \cite{pmlr-v70-guo17a}, ensemble temperate scaling \cite{pmlr-v119-zhang20k} and SOTA GNN-specific calibration methods, i.e., CaGCN \cite{NEURIPS2021_c7a9f13a}, GATS \cite{NEURIPS2022_5975754c} and CF-GNN\cite{NEURIPS2023_54a1495b}.
\subsubsection{Implementation.}
Our model and baselines are trained on Intel(R) Core(TM) i7-5820K CPU @ 3.30GHz, 64G RAM computing server, equipped with NVIDIA GTX TITAN X graphics cards. All hyper-parameters are chosen via random search. Details of hyper-parameter setting ranges are listed in Table \ref{tab:table_a2}.
% More details can be found in our available code.
\begin{table}[tb]
\centering
\begin{tabular}{cc}
\toprule
Param. & Value \\
\midrule
\(\lambda\)& \{1e-2, 1e-1, 1, 10\}\\ 
\(\tau\)& \{1e-2, 1e-1, 1, 10\}\\ 
\(\kappa\)& \{0, 1\} \\
GNN Layers & \{1, 2, 3, 4\} \\
GNN Hidden Dimension& \{16, 32, 64, 128, 256\} \\
Learning Rate & \{1e-1, 1e-2, 1e-3, 1e-4\} \\
\bottomrule
\end{tabular}
\caption{Hyper-parameters setting.}
\label{tab:table_a2}
\end{table}
\subsubsection{Metrics}
Marginal coverage and inefficiency are two commonly used metrics for measuring the performance of CP. Given the test set \(\mathcal{D}_{te}\), the marginal coverage metric is defined as follows:
\begin{equation}
    \textrm{Coverage}:=\frac{1}{|\mathcal{D}_{te}|} \sum_{i \in \mathcal{D}_{te}}{\delta[y_i \in C(x_i)]}
\end{equation}
where \(\delta[\cdot]\) is the indicator function, it is 1 when its argument is true and 0 otherwise. The coverage is empirically guaranteed when marginal coverage exceeds \(1-\alpha\). In cases where it exceeds this threshold, the results improve as they get closer to the target. The marginal coverage guarantee ensures that the output prediction sets for new test points provably include the true outcome with probability at least \(1-\alpha\). Then we focus on desirable prediction set size to enable further comparisons across CP methods. The inefficiency metric is defined by the size of the prediction set:
\begin{equation}
    \textrm{Ineff}:=\frac{1}{|\mathcal{D}_{te}|}\sum_{i \in \mathcal{D}_{te}}{| C(x_i)|}
\end{equation}
where smaller values indicate better performance.

\subsection{Results}

\subsubsection{Marginal Coverage Results. }
\begin{table*}[htb]
\centering
\begin{tabular}{cccccccc}
\toprule
Datasets& Temp. Scale. & Vector Scale. & Ensemble TS & CaGCN & GATS & CF-GNN & \textbf{RCP-GNN}\\
\midrule
Cora & 0.946\small{(.003)} &   0.944\small{(.004)}&  0.947\small{(.003)}&  0.939\small{(.005)}&  0.939\small{(.005)}&  \underline{0.952}\small{(.001)}& \underline{\textbf{0.950}}\small{(.002)}\\
DBLP & 0.920\small{(.009)}& 0.921\small{(.009)} & 0.920\small{(.008)}& 0.922\small{(.004)}& 0.921\small{(.004)}& \underline{0.952}\small{(.001)}& \underline{\textbf{0.950}}\small{(.001)}\\
CiteSeer & \underline{0.952}\small{(.004)}&  \underline{\textbf{0.951}}\small{(.004)}& \underline{0.953}\small{(.003)}& 0.949\small{(.005)}& \underline{\textbf{0.951}}\small{(.005)}& \underline{0.953}\small{(.001)}& \underline{\textbf{0.951}}\small{(.001)}\\
PubMed & 0.899\small{(.002)}&  0.899\small{(.003)}& 0.899\small{(.002)}& 0.898\small{(.003)} & 0.898\small{(.002)}& \underline{0.953}\small{(.001)}& \underline{\textbf{0.950}}\small{(.002)}\\
Computers & 0.929\small{(.002)}& 0.932\small{(.002)}& 0.930\small{(.002)}& 0.926\small{(.003)}& 0.925\small{(.002)} & \underline{0.952}\small{(.001)}& \underline{\textbf{0.951}}\small{(.001)}\\
Photo & \underline{0.962}\small{(.002)}& \underline{0.963}\small{(.002)}& \underline{0.964}\small{(.002)}& \underline{0.956}\small{(.002)}& \underline{0.957}\small{(.002)}& \underline{0.953}\small{(.001)} & \underline{\textbf{0.951}}\small{(.001)}\\

CS & \underline{0.957}\small{(.001)}& \underline{0.958}\small{(.001)}& \underline{0.958}\small{(.001)}& \underline{0.954}\small{(.003)}& \underline{0.957}\small{(.001)}& \underline{0.952}\small{(.001)} & \underline{\textbf{0.950}}\small{(.001)}\\

Physics & \underline{0.969}\small{(.000)}& \underline{0.969}\small{(.000)}& \underline{0.969}\small{(.000)}& \underline{0.968}\small{(.001)}& \underline{0.968}\small{(.000)}& \underline{0.952}\small{(.001)}& \underline{\textbf{0.950}}\small{(.001)} \\
\bottomrule
\end{tabular}
\caption{Empirical marginal coverage of different methods with \(\alpha = 0.05\). The result takes the average and standard deviation across 10 runs with 100 calib/test splits. Marked: \underline{Covered}, \textbf{Closest}.}
\label{table1_marginal_coverage}
\end{table*}
\begin{table*}[htb]
\centering
\begin{tabular}{cccccc}
\toprule
Methods & Cora & DBLP & CiteSeer & Computers & Photo  \\
\midrule
Temp. Scale. & 1.37&  1.19& 1.14& 1.31& 1.15\\
Vector Scale. & 1.36& 1.20& 1.15& 1.25& 1.13\\
Ensemble TS. & 1.37& 1.19& 1.14& 1.30& 1.15\\
CaGCN. & 1.41& \underline{1.18}& 1.19& \underline{1.22}& 1.14\\
GATS & \underline{1.33}& \underline{1.18}& 1.16& 1.28& \underline{1.12}\\
CF-GNN & 1.72& 1.23& \underline{0.99}& 1.81& 1.66\\
\midrule
RCP-GNN & \textbf{1.18}\small{(11.28\%\(\downarrow\))}& \textbf{1.17}\small{(0.85\%\(\downarrow\))}& \textbf{0.97}\small{(2.02\%\(\downarrow\))}& \textbf{1.20}\small{(1.64\%\(\downarrow\))}& \textbf{1.04}\small{(7.14\%\(\downarrow\))}\\
\bottomrule
\end{tabular}
\caption{Empirical inefficiency results of different methods across various datasets at test time with \(\alpha = 0.1\). Marked: \textbf{First.} \underline{Second.} The average inefficiency reduction relative to the best results of baselines in percentage is reported in parentheses.}

\label{table2_size}
\end{table*}
\begin{figure}[tb]
\centering
\includegraphics[width=1\columnwidth]{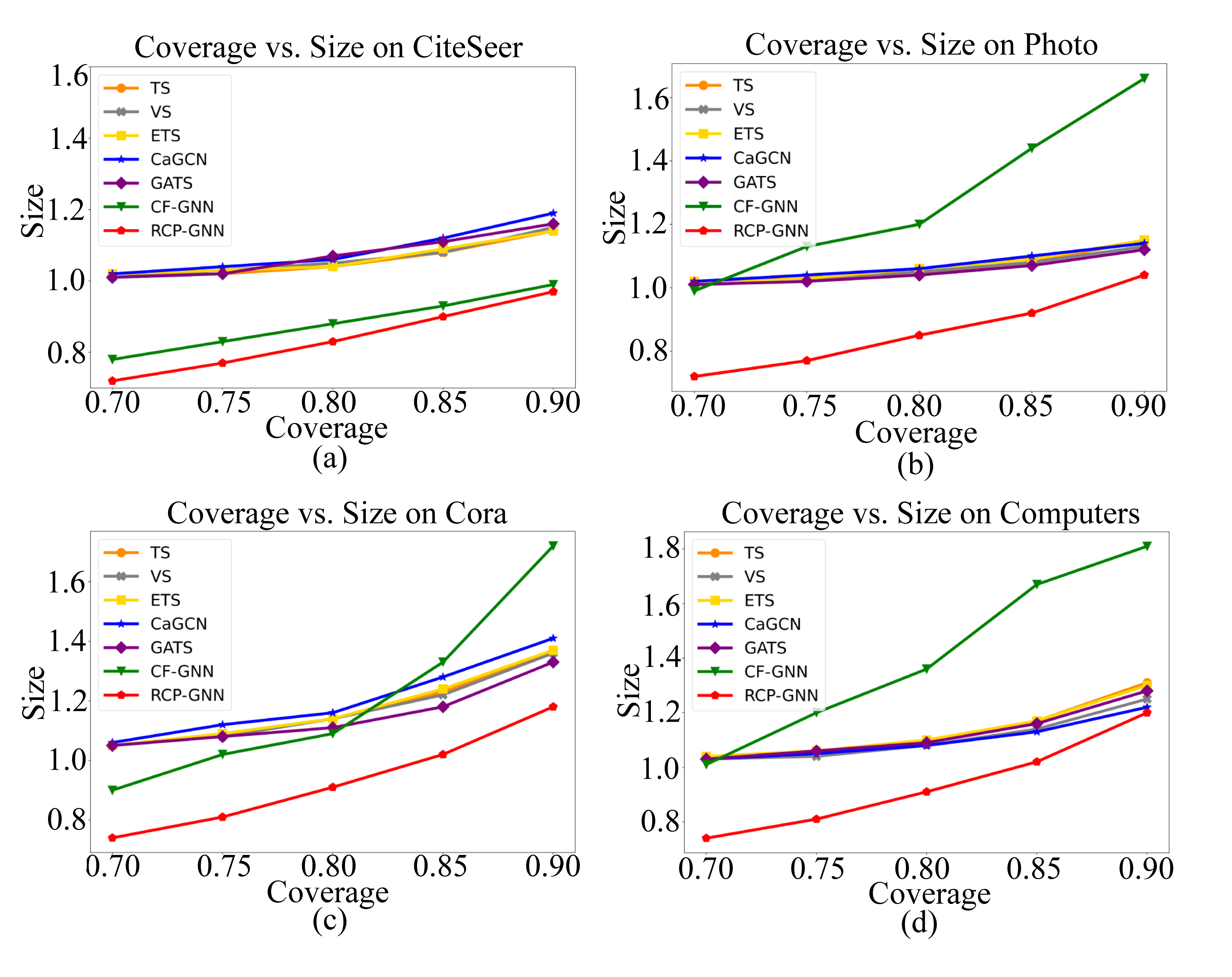} % Reduce the figure size so that it is slightly narrower than the column.
\caption{Results on different datasets. A lower curve means that the method can achieve the desired coverage using a smaller prediction set size.}
\label{fig2}
\end{figure}
\begin{figure}[tb]
\centering
\includegraphics[width=1\columnwidth]{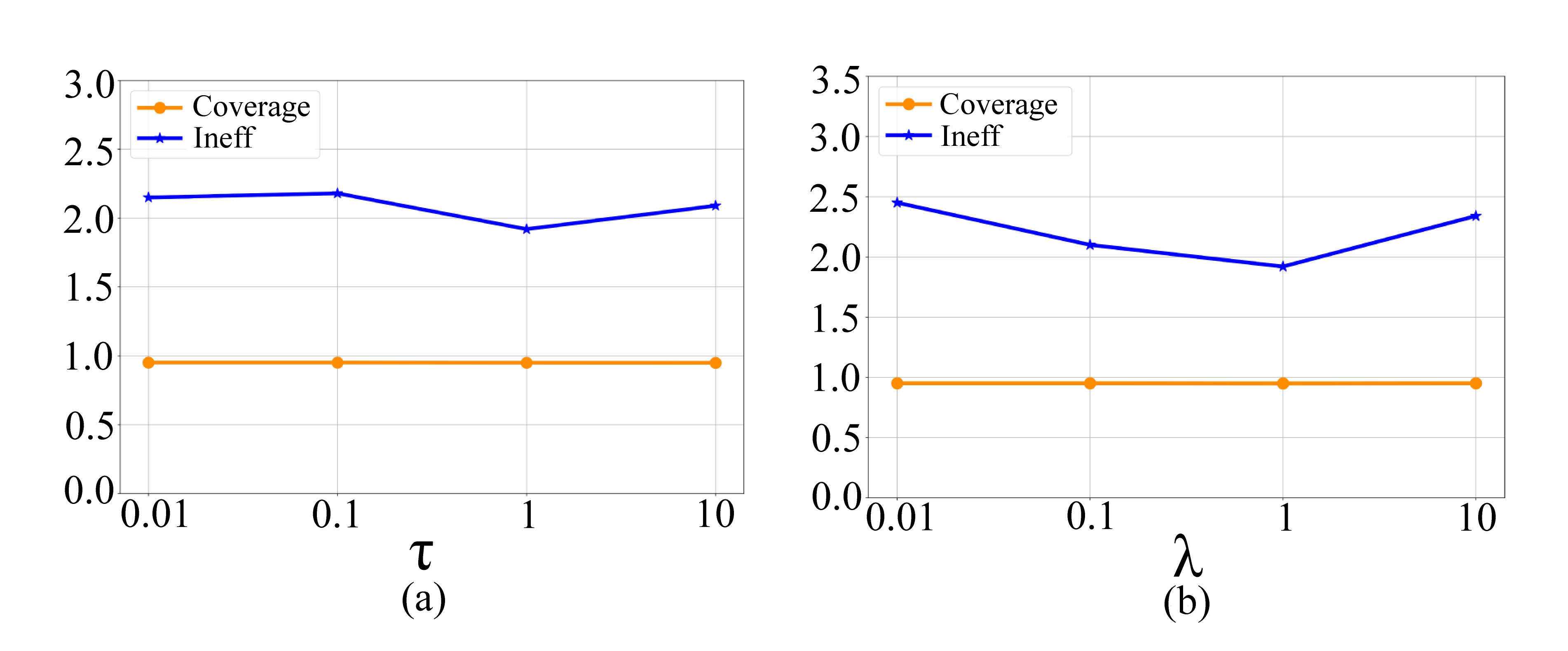} % Reduce the figure size so that it is slightly narrower than the column.
\caption{The coverage and inefficiency results with (a) \(\tau\) and (b) \(\lambda\) changes.}
\label{fig3}
\end{figure}
\begin{table*}[htb]
\centering
\begin{tabular}{c|c|c|c|c|c|c}
\toprule
\multirow{2}*{Methods} & \multicolumn{2}{c|}{Cora} & \multicolumn{2}{c|}{CiteSeer} & \multicolumn{2}{c}{Photo}\\
\cline{2-7} & Coverage & Size & Coverage & Size & Coverage & Size  \\
\midrule
RCP-THR & 0.954& \textbf{1.65}& 0.953& 1.57& 0.953& 1.26\\
RCP-APS & 0.952& 2.32& 0.952& 1.86& 0.953& 1.87\\
w/o Conf.Tr. & \textbf{0.950}& 2.08& \textbf{0.951}& 1.97& \textbf{0.951}& 2.00\\
\midrule
RCP-GNN & \textbf{0.950}& 1.92& \textbf{0.951}& \textbf{1.26}& \textbf{0.951}& \textbf{1.23}\\
\bottomrule
\end{tabular}
\caption{Empirical marginal coverage and inefficiency of variants at \(\alpha = 0.05\). Marked: \textbf{Best}.}
\label{table3_ablation_study}
\end{table*}

The marginal coverage of different methods are reported in Table \ref{table1_marginal_coverage}. All methods use the same pre-trained base model to avoid randomness. The results that achieves the target coverage are marked with underline. The most closest values among covered coverage are marked by bold font. Some observations are introduced as follows:

Temperate scaling, vector scaling and ensemble temperate scaling do not perform well because they lack to aware the topology information in graphs. Although CaGCN and GATS try to integrate topology information to per-node temperature scaling, their performance is still unsatisfactory because they only use CP as a post-training calibration step. Among SOTA methods, only CF-GNN reached target coverage on all datasets. It confirms that fixing prediction sets during training is helpful for reliable uncertainty estimates.

Among all methods, our model successfully reached target coverage on all datasets. Moreover, our model empirically achieves coverage rate closest to target. In summary, our model achieves superior empirical marginal coverage than existing methods.

\subsubsection{Inefficiency Results. }

In Table \ref{table2_size}, we summarize the inefficiency reductions of our methods in comparison to other baselines. It can be observed that our model achieve efficiency improvement across datasets with up to 11.28\% reduction in the prediction size.
We also empirical present the inefficiency of different methods on various tasks for \(\alpha\) ranging from 0.1 to 0.3 in Figure \ref{fig2}. Though CF-GNN try to reduce inefficiency through training with conformal prediction, it does not consistently improve inefficiency across all datasets. Specifically, on Amazon-Photo and Amazon-Computers, efficiency becomes even worse. Our RCP-GNN, in contrast, reduces inefficiency consistently.

The reason may be that RCP-GNN constructs and adjusts prediction sets based on ranking and probability of the labels, while CF-GNN only rely on assumptions about the model’s probabilities. Therefore, CF-GNN may not fully capture the model’s uncertainty, which hinders its performance. Additionally, RCP-GNN uses differentiable method to approximate ``hard'' sorting procedure, which helps to improve the performance and scalability of our framework.

In summary, our model can significantly reduce the inefficiency while maintaining satisfactory marginal coverage compared with other state-of-the-art methods.

\subsection{Ablation Study}

We conduct ablations in Table \ref{table3_ablation_study} to test main components in RCP-GNN. 1) \textit{RCP-THR.} It is a variant of RCP-GNN that using THR to compute the conformity scores. Since the THR conformity scores is naturally differentiable w.r.t. the model parameters \(\theta\) according to Eq. \ref{thr_v}, we only need to ensure the quantile function differentiable in the conformal training stage. 2) \textit{RCP-APS.} Similar to RCP-THR, this variant leverage APS to compute the conformity scores. The differentiable implementation closely follows the one for RANK outlined in Eq. \ref{Rank_diff_score}:
\begin{equation}
    V (x_i, y_i) = \sum_{k=1}^{|\mathcal{Y}|}\textrm{sigmoid}(\frac{\tilde{\mu}_{y_i}(x_i) - \tilde{\mu}_{k}(x_i)}{\tau}) \tilde{\mu}_{k}(x_i)
\end{equation}
3) \textit{w/o Conf.Tr.} In order to figure out the power of conformal training step, we remove the conformity loss and replace it with standard prediction loss.
Compared with RCP-THR and RCP-APS, our model can achieve pre-defined marginal coverage with satisfactory inefficiency reduction,
% We first experimental demonstrate our model can improve efficiency of THR, APS and RAPS applied to a baseline model. 
which demonstrates that the rank-based conformal prediction component is performance-critical to ensure valid coverage guarantees while simultaneously enhancing efficiency. Compared with w/o Conf.Tr., our model achieves consistent efficiency improvement, which demonstrates that prediction sets can be optimized along with conformal training.

\subsection{Hyper-Parameter Sensitivity}

We also conduct experiments for major hyper-parameters of our model to test the robustness of RCP-GNN. In concrete, the hyper-parameter temperature is changed from 0.01 to 10 and the results show that our model is not sensitivity to the temperature. And we select the median value 1 for its relatively better performance. The hyper-parameter \(\lambda\) in Eq. \ref{eq_loss} is used to balance the prediction loss and the conformity loss. We report the converge and inefficiency results as \(\lambda\) changes from 0.01 to 10 and we can observe that a proper weight of conformity loss can help to inefficiency reduction.

\section{Related Works}
 
\subsubsection{Uncertainty Quantification in Deep Learning.}
 It is important for trustworthy modern deep learning models to mitigate overconfidence \cite{Wang2020WisdomOT, Slossberg2020OnCO, 10.5555/3327345.3327458}. Uncertainty quantification (UQ), which aims to construct model-agnostic uncertain estimates, have great potential in many high-stakes applications \cite{ABDAR2021243, Gupta2020CalibrationON, pmlr-v70-guo17a, NEURIPS2019_8ca01ea9, pmlr-v119-zhang20k}. Most of existing UQ methods rely on the i.i.d assumption. Thus make them be not easily adopt to inter-dependency graph-structure data. Some network principle-based UQ methods  \cite{NEURIPS2021_c7a9f13a, NEURIPS2022_5975754c} designing for GNNs have been proposed in recent years. However, these UQ methods fail to achieve valid coverage guarantee.

\subsubsection{Standard Conformal Prediction.}
Conformal prediction (CP) is early proposed on \cite{Vovk2005AlgorithmicLI}.
Compared with other CP framework, e.g., cross-validation \cite{vovk2012crossconformalpredictors} or jackknife \cite{barber2020predictiveinferencejackknife}, most approaches follow a split CP method \cite{Lei2013ACP}, where a held-out calibration set is necessary. For the reason that it defines faster and more scalable CP algorithms. However, it sacrifices statistical efficiency. 

Different variants of CP are struggled to the balance between statistical and computational efficiency.
Some contributions made in conformity score function have been explored \cite{Sadinle_2018, Angelopoulos2020UncertaintySF, NEURIPS2020_244edd7e} for inefficiency reduction. 
Other studies \cite{Bates2021DistributionFreeRP, yang2024selectionaggregationconformalprediction} have explored in the context of ensembles to obtain smaller confidence sets while avoiding to sacrifice the obtained empirical coverage.
But these methods do not solve the major limitation of CP methods: the model is independent, leaving CP little to no control over the prediction sets \cite{NIPS2012_cfbce4c1}. Recently, the work of \cite{Bellotti2021OptimizedCC} and \cite{stutz2022learning} try to better integrated CP into deep learning models by simulating CP during training to make full use of CP benefits. For GNNs, how to define a trainable calibration step still remains an open space for exploration.

\subsubsection{Conformal Prediction for Graphs.}
Some efforts have been done for CP to GNNs. The work of \cite{Wijegunawardana2020NodeCW} adapted CP for node classification to achieve bounded error, and \cite{pmlr-v202-clarkson23a} adapted weighted exchangeability without any lowerbound on the coverage. Furthermore, the assumption for a valid guarantee that the exchangeability between the calibration set and the test set is proved by \cite{pmlr-v202-h-zargarbashi23a, NEURIPS2023_54a1495b}, which makes CP applicable to transduction node classification tasks.
Different from these works, we leverage the rank of prediction probabilities of nodes to reduce its miscalibration. We also provide its differentiable variant for calibration during training to make prediction sets become aware of network topology information.

\section{Conclusion}

In this work, we extend CP to GNNs by proposing a trainable rank-based CP framework for marginal coverage guaranteed and inefficiency reduction. In future work we will focus on more tasks like link prediction, and extensions to graph-based applications such as molecular prediction and recommendation systems.

\section*{Acknowledgements}
The work described in this paper was partially supported by grants from City University of Hong Kong (Project No. 9610639, 6000864).

\bibliography{aaai25}

\begin{thebibliography}{39}
\providecommand{\natexlab}[1]{#1}

\bibitem[{Abdar et~al.(2021)Abdar, Pourpanah, Hussain, Rezazadegan, Liu, Ghavamzadeh, Fieguth, Cao, Khosravi, Acharya, Makarenkov, and Nahavandi}]{ABDAR2021243}
Abdar, M.; Pourpanah, F.; Hussain, S.; Rezazadegan, D.; Liu, L.; Ghavamzadeh, M.; Fieguth, P.; Cao, X.; Khosravi, A.; Acharya, U.~R.; Makarenkov, V.; and Nahavandi, S. 2021.
\newblock A review of uncertainty quantification in deep learning: Techniques, applications and challenges.
\newblock \emph{Information Fusion}, 76: 243--297.

\bibitem[{Angelopoulos et~al.(2021)Angelopoulos, Bates, Jordan, and Malik}]{Angelopoulos2020UncertaintySF}
Angelopoulos, A.~N.; Bates, S.; Jordan, M.; and Malik, J. 2021.
\newblock Uncertainty Sets for Image Classifiers using Conformal Prediction.
\newblock In \emph{International Conference on Learning Representations}.

\bibitem[{Barber et~al.(2021)Barber, Candes, Ramdas, and Tibshirani}]{barber2020predictiveinferencejackknife}
Barber, R.~F.; Candes, E.~J.; Ramdas, A.; and Tibshirani, R.~J. 2021.
\newblock Predictive inference with the jackknife+.
\newblock \emph{The Annals of Statistics}, 49(1): 486--507.

\bibitem[{Bates et~al.(2021)Bates, Angelopoulos, Lei, Malik, and Jordan}]{Bates2021DistributionFreeRP}
Bates, S.; Angelopoulos, A.; Lei, L.; Malik, J.; and Jordan, M. 2021.
\newblock Distribution-free, risk-controlling prediction sets.
\newblock \emph{Journal of the ACM (JACM)}, 68(6): 1--34.

\bibitem[{Bellotti(2021)}]{Bellotti2021OptimizedCC}
Bellotti, A. 2021.
\newblock Optimized conformal classification using gradient descent approximation.
\newblock arXiv:2105.11255.

\bibitem[{Blondel et~al.(2020)Blondel, Teboul, Berthet, and Djolonga}]{pmlr-v119-blondel20a}
Blondel, M.; Teboul, O.; Berthet, Q.; and Djolonga, J. 2020.
\newblock Fast Differentiable Sorting and Ranking.
\newblock In \emph{Proceedings of the 37th International Conference on Machine Learning}, volume 119, 950--959. PMLR.

\bibitem[{Chernozhukov, Fern{\'a}ndez-Val, and Galichon(2007)}]{Chernozhukov2007QUANTILEAP}
Chernozhukov, V.; Fern{\'a}ndez-Val, I.; and Galichon, A. 2007.
\newblock Quantile and probability curves without crossing.
\newblock \emph{Econometrica}, 78: 1093--1125.

\bibitem[{Clarkson(2023)}]{pmlr-v202-clarkson23a}
Clarkson, J. 2023.
\newblock Distribution Free Prediction Sets for Node Classification.
\newblock In \emph{Proceedings of the 40th International Conference on Machine Learning}, volume 202, 6268--6278. PMLR.

\bibitem[{Guo et~al.(2017)Guo, Pleiss, Sun, and Weinberger}]{pmlr-v70-guo17a}
Guo, C.; Pleiss, G.; Sun, Y.; and Weinberger, K.~Q. 2017.
\newblock On Calibration of Modern Neural Networks.
\newblock In \emph{Proceedings of the 34th International Conference on Machine Learning}, volume~70, 1321--1330. PMLR.

\bibitem[{Gupta et~al.(2021)Gupta, Rahimi, Ajanthan, Mensink, Sminchisescu, and Hartley}]{Gupta2020CalibrationON}
Gupta, K.; Rahimi, A.; Ajanthan, T.; Mensink, T.; Sminchisescu, C.; and Hartley, R. 2021.
\newblock Calibration of Neural Networks using Splines.
\newblock In \emph{International Conference on Learning Representations}.

\bibitem[{Guzm\'{a}n-rivera, Batra, and Kohli(2012)}]{NIPS2012_cfbce4c1}
Guzm\'{a}n-rivera, A.; Batra, D.; and Kohli, P. 2012.
\newblock Multiple Choice Learning: Learning to Produce Multiple Structured Outputs.
\newblock In \emph{Advances in Neural Information Processing Systems}, volume~25. Curran Associates, Inc.

\bibitem[{H.~Zargarbashi, Antonelli, and Bojchevski(2023)}]{pmlr-v202-h-zargarbashi23a}
H.~Zargarbashi, S.; Antonelli, S.; and Bojchevski, A. 2023.
\newblock Conformal Prediction Sets for Graph Neural Networks.
\newblock In \emph{Proceedings of the 40th International Conference on Machine Learning}, volume 202, 12292--12318. PMLR.

\bibitem[{Hsu et~al.(2022)Hsu, Shen, Tomani, and Cremers}]{NEURIPS2022_5975754c}
Hsu, H. H.-H.; Shen, Y.; Tomani, C.; and Cremers, D. 2022.
\newblock What Makes Graph Neural Networks Miscalibrated?
\newblock In \emph{Advances in Neural Information Processing Systems}, volume~35, 13775--13786. Curran Associates, Inc.

\bibitem[{Huang et~al.(2023)Huang, Jin, Candes, and Leskovec}]{NEURIPS2023_54a1495b}
Huang, K.; Jin, Y.; Candes, E.; and Leskovec, J. 2023.
\newblock Uncertainty Quantification over Graph with Conformalized Graph Neural Networks.
\newblock In \emph{Advances in Neural Information Processing Systems}, volume~36, 26699--26721. Curran Associates, Inc.

\bibitem[{Jiang et~al.(2018)Jiang, Kim, Guan, and Gupta}]{10.5555/3327345.3327458}
Jiang, H.; Kim, B.; Guan, M.~Y.; and Gupta, M. 2018.
\newblock To trust or not to trust a classifier.
\newblock In \emph{Proceedings of the 32nd International Conference on Neural Information Processing Systems}, NIPS'18, 5546–5557. Red Hook, NY, USA: Curran Associates Inc.

\bibitem[{Kull et~al.(2019)Kull, Perello~Nieto, K\"{a}ngsepp, Silva~Filho, Song, and Flach}]{NEURIPS2019_8ca01ea9}
Kull, M.; Perello~Nieto, M.; K\"{a}ngsepp, M.; Silva~Filho, T.; Song, H.; and Flach, P. 2019.
\newblock Beyond temperature scaling: Obtaining well-calibrated multi-class probabilities with Dirichlet calibration.
\newblock In \emph{Advances in Neural Information Processing Systems}, volume~32. Curran Associates, Inc.

\bibitem[{Lam et~al.(2023)Lam, Sanchez-Gonzalez, Willson, Wirnsberger, Fortunato, Alet, Ravuri, Ewalds, Eaton-Rosen, Hu, Merose, Hoyer, Holland, Vinyals, Stott, Pritzel, Mohamed, and Battaglia}]{doi:10.1126/science.adi2336}
Lam, R.; Sanchez-Gonzalez, A.; Willson, M.; Wirnsberger, P.; Fortunato, M.; Alet, F.; Ravuri, S.; Ewalds, T.; Eaton-Rosen, Z.; Hu, W.; Merose, A.; Hoyer, S.; Holland, G.; Vinyals, O.; Stott, J.; Pritzel, A.; Mohamed, S.; and Battaglia, P. 2023.
\newblock Learning skillful medium-range global weather forecasting.
\newblock \emph{Science}, 382(6677): 1416--1421.

\bibitem[{Lei, Rinaldo, and Wasserman(2013)}]{Lei2013ACP}
Lei, J.; Rinaldo, A.; and Wasserman, L.~A. 2013.
\newblock A conformal prediction approach to explore functional data.
\newblock \emph{Annals of Mathematics and Artificial Intelligence}, 74: 29 -- 43.

\bibitem[{Li, Huang, and Zitnik(2022)}]{GRLBH}
Li, M.~M.; Huang, K.; and Zitnik, M. 2022.
\newblock Graph representation learning in biomedicine and healthcare.
\newblock \emph{Nature Biomedical Engineering}, 6(12): 1353--1369.

\bibitem[{Lunde(2023)}]{lunde2023validityconformalpredictionnetwork}
Lunde, R. 2023.
\newblock On the Validity of Conformal Prediction for Network Data Under Non-Uniform Sampling.
\newblock arXiv:2306.07252.

\bibitem[{Lunde, Levina, and Zhu(2023)}]{lunde2023conformalpredictionnetworkassistedregression}
Lunde, R.; Levina, E.; and Zhu, J. 2023.
\newblock Conformal Prediction for Network-Assisted Regression.
\newblock arXiv:2302.10095.

\bibitem[{Luo and Zhou(2024)}]{luo2024trustworthy}
Luo, R.; and Zhou, Z. 2024.
\newblock Trustworthy Classification through Rank-Based Conformal Prediction Sets.
\newblock arXiv:2407.04407.

\bibitem[{Marandon(2024)}]{marandon2024conformallinkpredictionfalse}
Marandon, A. 2024.
\newblock Conformal link prediction for false discovery rate control.
\newblock \emph{TEST}, 1--22.

\bibitem[{Romano, Sesia, and Candes(2020)}]{NEURIPS2020_244edd7e}
Romano, Y.; Sesia, M.; and Candes, E. 2020.
\newblock Classification with Valid and Adaptive Coverage.
\newblock In \emph{Advances in Neural Information Processing Systems}, volume~33, 3581--3591. Curran Associates, Inc.

\bibitem[{Sadinle, Lei, and Wasserman(2019)}]{Sadinle_2018}
Sadinle, M.; Lei, J.; and Wasserman, L. 2019.
\newblock Least ambiguous set-valued classifiers with bounded error levels.
\newblock \emph{Journal of the American Statistical Association}, 114(525): 223--234.

\bibitem[{Shchur et~al.(2019)Shchur, Mumme, Bojchevski, and Günnemann}]{DBLP:journals/corr/abs-1811-05868}
Shchur, O.; Mumme, M.; Bojchevski, A.; and Günnemann, S. 2019.
\newblock Pitfalls of Graph Neural Network Evaluation.
\newblock arXiv:1811.05868.

\bibitem[{Slossberg et~al.(2022)Slossberg, Anschel, Markovitz, Litman, Aberdam, Tsiper, Mazor, Wu, and Manmatha}]{Slossberg2020OnCO}
Slossberg, R.; Anschel, O.; Markovitz, A.; Litman, R.; Aberdam, A.; Tsiper, S.; Mazor, S.; Wu, J.; and Manmatha, R. 2022.
\newblock On calibration of scene-text recognition models.
\newblock In \emph{European Conference on Computer Vision}, 263--279. Springer.

\bibitem[{Stadler et~al.(2021)Stadler, Charpentier, Geisler, Z\"{u}gner, and G\"{u}nnemann}]{NEURIPS2021_95b431e5}
Stadler, M.; Charpentier, B.; Geisler, S.; Z\"{u}gner, D.; and G\"{u}nnemann, S. 2021.
\newblock Graph Posterior Network: Bayesian Predictive Uncertainty for Node Classification.
\newblock In \emph{Advances in Neural Information Processing Systems}, volume~34, 18033--18048. Curran Associates, Inc.

\bibitem[{Stutz et~al.(2022)Stutz, Dvijotham, Cemgil, and Doucet}]{stutz2022learning}
Stutz, D.; Dvijotham, K.~D.; Cemgil, A.~T.; and Doucet, A. 2022.
\newblock Learning Optimal Conformal Classifiers.
\newblock In \emph{International Conference on Learning Representations}.

\bibitem[{Vovk(2015)}]{vovk2012crossconformalpredictors}
Vovk, V. 2015.
\newblock Cross-conformal predictors.
\newblock \emph{Annals of Mathematics and Artificial Intelligence}, 74: 9--28.

\bibitem[{Vovk, Gammerman, and Shafer(2005)}]{Vovk2005AlgorithmicLI}
Vovk, V.; Gammerman, A.; and Shafer, G. 2005.
\newblock \emph{Algorithmic learning in a random world}.
\newblock Springer.

\bibitem[{Wang et~al.(2024)Wang, Liu, Liu, Wang, Medya, and Yu}]{wang2024uncertainty}
Wang, F.; Liu, Y.; Liu, K.; Wang, Y.; Medya, S.; and Yu, P.~S. 2024.
\newblock Uncertainty in Graph Neural Networks: A Survey.
\newblock arXiv:2403.07185.

\bibitem[{Wang et~al.(2020)Wang, Ghosh, Gonzalez~Diaz, Farahat, Alam, Gupta, Chen, and Marathe}]{Wang2020WisdomOT}
Wang, L.; Ghosh, D.; Gonzalez~Diaz, M.; Farahat, A.; Alam, M.; Gupta, C.; Chen, J.; and Marathe, M. 2020.
\newblock Wisdom of the Ensemble: Improving Consistency of Deep Learning Models.
\newblock In \emph{Advances in Neural Information Processing Systems}, volume~33, 19750--19761. Curran Associates, Inc.

\bibitem[{Wang et~al.(2021)Wang, Liu, Shi, and Yang}]{NEURIPS2021_c7a9f13a}
Wang, X.; Liu, H.; Shi, C.; and Yang, C. 2021.
\newblock Be Confident! Towards Trustworthy Graph Neural Networks via Confidence Calibration.
\newblock In \emph{Advances in Neural Information Processing Systems}, volume~34, 23768--23779. Curran Associates, Inc.

\bibitem[{Wijegunawardana, Gera, and Soundarajan(2020)}]{Wijegunawardana2020NodeCW}
Wijegunawardana, P.; Gera, R.; and Soundarajan, S. 2020.
\newblock Node Classification with Bounded Error Rates.
\newblock In \emph{Complex Networks XI: Proceedings of the 11th Conference on Complex Networks CompleNet 2020}, 26--38. Springer.

\bibitem[{Wu et~al.(2022)Wu, Sun, Zhang, Xie, and Cui}]{10.1145/3535101}
Wu, S.; Sun, F.; Zhang, W.; Xie, X.; and Cui, B. 2022.
\newblock Graph neural networks in recommender systems: a survey.
\newblock \emph{ACM Computing Surveys}, 55(5): 1--37.

\bibitem[{Yang and Kuchibhotla(2024)}]{yang2024selectionaggregationconformalprediction}
Yang, Y.; and Kuchibhotla, A.~K. 2024.
\newblock Selection and Aggregation of Conformal Prediction Sets.
\newblock \emph{Journal of the American Statistical Association}, 1--13.

\bibitem[{Yang, Cohen, and Salakhudinov(2016)}]{pmlr-v48-yanga16}
Yang, Z.; Cohen, W.; and Salakhudinov, R. 2016.
\newblock Revisiting Semi-Supervised Learning with Graph Embeddings.
\newblock In \emph{Proceedings of The 33rd International Conference on Machine Learning}, volume~48, 40--48. PMLR.

\bibitem[{Zhang, Kailkhura, and Han(2020)}]{pmlr-v119-zhang20k}
Zhang, J.; Kailkhura, B.; and Han, T. Y.-J. 2020.
\newblock Mix-n-Match : Ensemble and Compositional Methods for Uncertainty Calibration in Deep Learning.
\newblock In \emph{Proceedings of the 37th International Conference on Machine Learning}, volume 119, 11117--11128. PMLR.

\end{thebibliography}

\end{document}